\documentclass[runningheads]{llncs}
\usepackage[dvipdfmx]{graphicx}
\usepackage{graphicx}
\usepackage{comment}
\usepackage{amsmath,amssymb} 
\usepackage{color}

\begin{document}
\pagestyle{headings}
\mainmatter

\title{Captioning Images with Novel Objects via \\Online Vocabulary Expansion} 


%
\author{Mikihiro Tanaka\inst{1} \and Tatsuya Harada\inst{1,2}}
\authorrunning{M. Tanaka et al.}
%

\institute{${}^\text{1}$The University of Tokyo, ${}^\text{2}$RIKEN}
\maketitle

\begin{abstract}
In this study, we introduce a low cost method for generating descriptions from images containing novel objects. Generally, constructing a model, which can explain images with novel objects, is costly because of the following: (1) collecting a large amount of data for each category, and (2) retraining the entire system. If humans see a small number of novel objects, they are able to estimate their properties by associating their appearance with known objects. Accordingly, we propose a method that can explain images with novel objects without retraining using the word embeddings of the objects estimated from only a small number of image features of the objects. The method can be integrated with general image-captioning models. The experimental results show the effectiveness of our approach.
\end{abstract}

\section{Introduction}
Image captioning is the task of generating a natural-language sentence that summarizes an input image, and plays an important role in intelligent systems for supporting and interacting with humans. The applications include automatic indexing for images uploaded on the Web, assisting the lives of people with visual disabilities by explaining to them what is in front of their eyes, and communication between humans and agents~\cite{Mao2016,Yu2017,Das2017}. In our lives, there is a wide array of various objects including new products such as drones and robots. For the above applications, an ideal system should be designed to describe images with the long-tailed and open-ended distribution of objects in the real world~\cite{Liu2019}.

However, in general image captioning, a system can describe newly input images by constructing an image-caption paired dataset first and learning the correspondence between images and sentences. Thus, the system cannot describe images with objects that are not included in the dataset. Here, we define \textit{objects that are not in image-caption paired datasets} as \textit{novel objects}. In general, to describe images with novel objects, creating a dataset with manual descriptions for images containing these objects and training a whole system are needed. However, (1) the data collection cost is high because, in addition to text annotation, collecting images itself might be difficult for some categories, and (2) computation cost is high for training whole systems. Even if image-caption-paired datasets are not available, the names of novel objects (e.g., image tags) are easier to obtain as they may be available as images on the Web or can be estimated by image taggers~\cite{Chen2013,Zhang2016}, and the names of novel objects in images are at least required for describing images with novel objects. Thus, we focus on describing images that include novel objects with tags during inference.

In recent years, some studies have been conducted on generating sentences from images with novel objects by utilizing external knowledge from (A) image classification/object detection datasets and (B) text-corpus such as Wikipedia. In \cite{Anderson_2017},
a method was proposed to describe images that include novel objects with tags under the vocabulary in (B) by employing word embeddings pretrained on (B). However, if a category is not present in (B), we must first collect a large number of sentences related to that category and then retrain, thereby incurring significant costs. \cite{Lu2018,Wu_2018,Feng_2020} aimed to describe images with novel objects that are included in (A) by leveraging an object detector where (B) was not available. However, if some objects are not even in (A), constructing an object detector is needed, which requires higher costs compared to obtaining image tags.

\begin{figure}[t]
\begin{center}
   \includegraphics[width=0.9\linewidth]{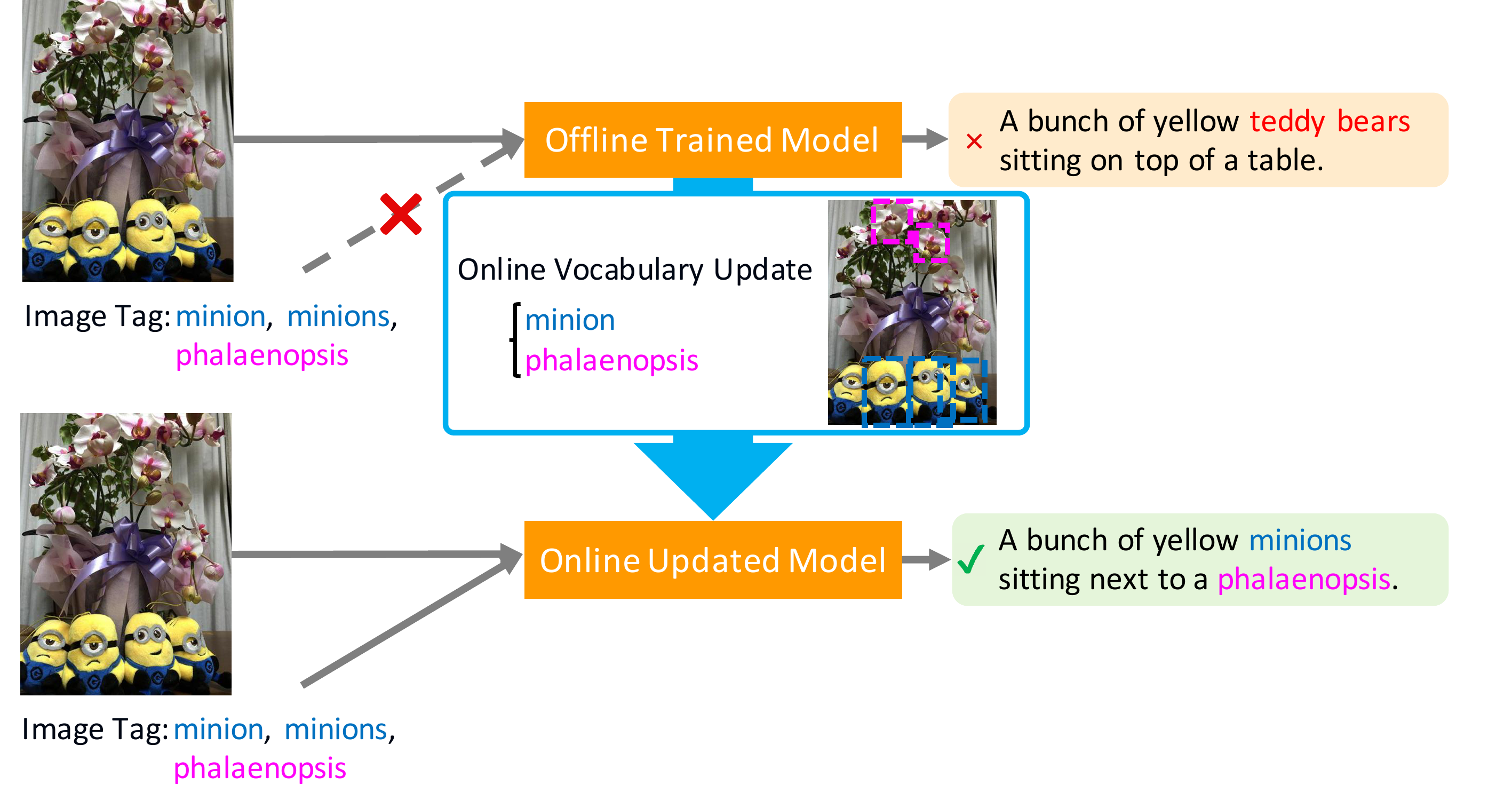}
\end{center}
\vspace{-15pt}
   \caption{Example of a sentence generated using our method. The bottom sentence is generated after online vocabulary update. The image-captioning model has no knowledge of a \textit{minion} and a \textit{phalaenopsis}, i.e., neither their images nor texts are seen during training. Bounding box annotations depicted in the figure are required only when the vocabulary is updated. After the vocabulary update, images with tags that include a \textit{minion} and a \textit{phalaenopsis} can be described without additional annotations.}
\label{fig:few_theme}
\end{figure}

We work on describing images that include novel objects with tags by a model trained on an image-caption paired dataset. To alleviate the issues arising in (1) and (2), we expand the vocabulary of this model using a small amount of annotations without requiring retraining. If a human sees a small number of images of novel objects, their properties can be estimated by comparing them with known objects with similar appearances. Estimating the properties of novel objects from visual information can also be a great clue for machines. Therefore, in this study, we propose a method that can explain images with novel objects by estimating the word embeddings of the objects using only a small number of image features of the objects. In this method, bounding box annotations of novel objects are required only when their word embeddings are estimated for the vocabulary expansion. Although some existing studies on different tasks use a similar way to get word embeddings~\cite{Frome2013,Norouzi2014}, they need large text-corpus and our method does not.

Fig.~\ref{fig:few_theme} shows an example sentence generated using our method. Although a \textit{minion} and a \textit{phalaenopsis} are not given in images or texts during training, the model can describe images with those using estimated word embeddings from image features in the bounding boxes without retraining. Therefore, it becomes easier to build a system that can explain images with various real-world objects by easing the problem (1) and (2). In this study, image tags are given by using ground-truths (GT). Image captioning that does not depend on the recognition performance of novel objects can be evaluated.

The contributions of this paper are as follows:
\begin{itemize}
 \setlength{\parskip}{0cm} 
 \setlength{\itemsep}{0cm} 
 \item We propose a novel task of describing images that include novel objects with tags by using a small amount of annotations without retraining in order to lower data collection costs and computation costs.
 \item We propose a method to solve this problem by using word embeddings of novel objects estimated from a small number of image features of the objects, and experimental results showed the effectiveness of our approach.
\end{itemize}

\section{Related Works}
First, we introduce image captioning. Second, we explain the captioning of images with novel objects. Finally, we summarize the difference between our study and the existing ones.

\noindent{\textbf{Image captioning.}} An encoder-decoder model that extracts features from images and decodes them by natural-language is commonly used. Many recent studies have adopted a deep neural network, namely, the deep convolutional neural network (CNN) for the encoder part and long short-term memory (LSTM) for the decoder part~\cite{Lu2017,Xu2015,You2016,Rennie2017,Anderson2018,Lu2018,Yao2018}. In addition, instead of using one LSTM, researches have been conducted to improve the sentence-generation performance using a mechanism called attention, which uses the local information of various image regions for each word to be generated. Models that use a CNN~\cite{Aneja2018} or a Transformer~\cite{Sharma2018} as decoders have also been proposed. In these models, a sentence is generated by repeatedly inputting the features of the output word, and output word is decided by calculating the similarity between the output from the decoder and the word embeddings trained on image-caption-paired dataset. Therefore, although some objects, which are not included in the training dataset, can be recognized, the model can not generate a sentence using those words. 

\noindent{\textbf{Captioning images with novel objects.}} In the previous researches on image captioning, image-caption-paired datasets such as MSCOCO~\cite{Lin2014}, Flickr 8K~\cite{Hodosh2013}, and Flickr 30K~\cite {Young2014}, etc. were used, and only images with objects included in these dataset were studied. However, in the real world, various objects exist, and, therefore, research works on captioning images with these various objects have been conducted in the recent years~\cite{Hendricks_2016,Yao_2017,Anderson_2017,Lu2018,Wu_2018,Venugopalan_2017,Anderson_2018,Mogadala2017,Agrawal_2019,Li_2019,Feng_2020}. For alternative sources for obtaining the knowledge of novel objects, many studies use (A) image-classification/object-detection datasets such as ImageNet~\cite{imagenet_cvpr09} and Open Image~\cite{openimages}, and (B) Web-based large-scale text-corpus such as Wikipedia.  \cite{Demirel_2019} conducted research in a situation wherein (A) could not be obtained, but only (B) was available. They utilized a zero-shot detector using the features of words learned from the text-corpus to generate caption with objects that appear in (B). Furthermore, \cite{Feng_2020} utilized an object detector as (A) to generate captions with objects included in (A) without using (B). They first generated sentences using the nouns of the known objects, and replaced them with the nouns of novel objects in the second stage. Similar to our research, there exists a research on captioning images using image tags~\cite{Anderson_2017} during inference, and it can be used in a situation wherein (A) cannot be obtained but (B) is available. They proposed a search algorithm called constrained beam search (CBS), which could enable the model to generate unseen words by using the word embeddings, Glove~\cite{Pennington2014} pretrained on text-corpus. \cite{Anderson_2018} regarded the sentences generated using CBS as GT captions, and improved the performance by proposing an iterative algorithm, which was inspired by expectation--maximization (EM)~\cite{Dempster1977}.

\noindent{\textbf{Summary.}} In this study, we examine captioning images with tags as \cite{Anderson_2017}. Furthermore, in our setting, (A) and (B) are not available during training, but a small number of (A) is available online, which is easier to be applied to real-world applications than in existing settings. We focus on tagged images because it is easier to construct image-tag paired datasets for training image taggers compared with other tasks such as object detection; in addition, many situations exist where tags can be automatically obtained. In the situation of \cite{Anderson_2017}, large text-corpus of novel objects were available and the research could not be adapted if the word embeddings of novel objects were not known during training. However, in this study, we focus on the fact that humans can estimate how to describe novel objects using their appearance information. Accordingly, we propose a method that can explain images with novel objects without undergoing retraining with the estimated word embeddings of the objects using only a small number of the image features of the objects.

\section{Proposed System}
In this study, a description is generated from an input image that include novel objects with tags. We propose a method that can be integrated with common image-captioning models, which sequentially output words using decoders such as LSTM, CNN, and Transformer. The following functions are required to construct a system that can explain images with various real-world objects:

\begin{itemize}
 \setlength{\parskip}{0cm} 
 \setlength{\itemsep}{0cm} 
 \item[(A)] Low data-collection cost; large amount of data is not required for novel categories.
 \item[(B)] Low computation cost; novel categories can be added with low training cost.
 \item[(C)] Maintain the performance of known categories.
\end{itemize}

To describe images with novel objects by a model that sequentially generates words, word embeddings of those categories are required. In this study, we propose a method to expand the vocabulary by estimating the word embeddings from the images of novel objects on the basis of the fact that humans can estimate the property of the novel objects by their appearances. We design a model that can add novel categories by using a small number of images without retraining, so that the required functions (A) and (B) can be satisfied. Generating a sentence $S$ for an image $I$ with novel objects $C_{novel}$ is expressed as follows:

\begin{eqnarray}
P(S|I)=P(C_{novel}|I)P(S|I,C_{novel})
\end{eqnarray}

When no novel object is present, the aforementioned equation means normal image captioning $P(S|I)$, and we build a system that satisfies the function (C) by not changing the word embeddings of the known categories when adding novel categories. The term $P(C_{novel}|I)$ denotes a problem of recognizing whether there exist novel objects in an image. In this study, we focus on $P(S|I,C_ {novel})$, which is a part of captioning images with novel objects, and we regarded that $P(C_{novel}|I)$ is given as image tags. 

First, we explain the manner to expand the vocabulary. Next, we introduce the procedure to generate captions from images with novel objects using the expanded vocabulary.

\subsection{Vocabulary Expansion}
\label{sec:ex}

\begin{figure*}[t!]
\centering
  \includegraphics[width=\linewidth]{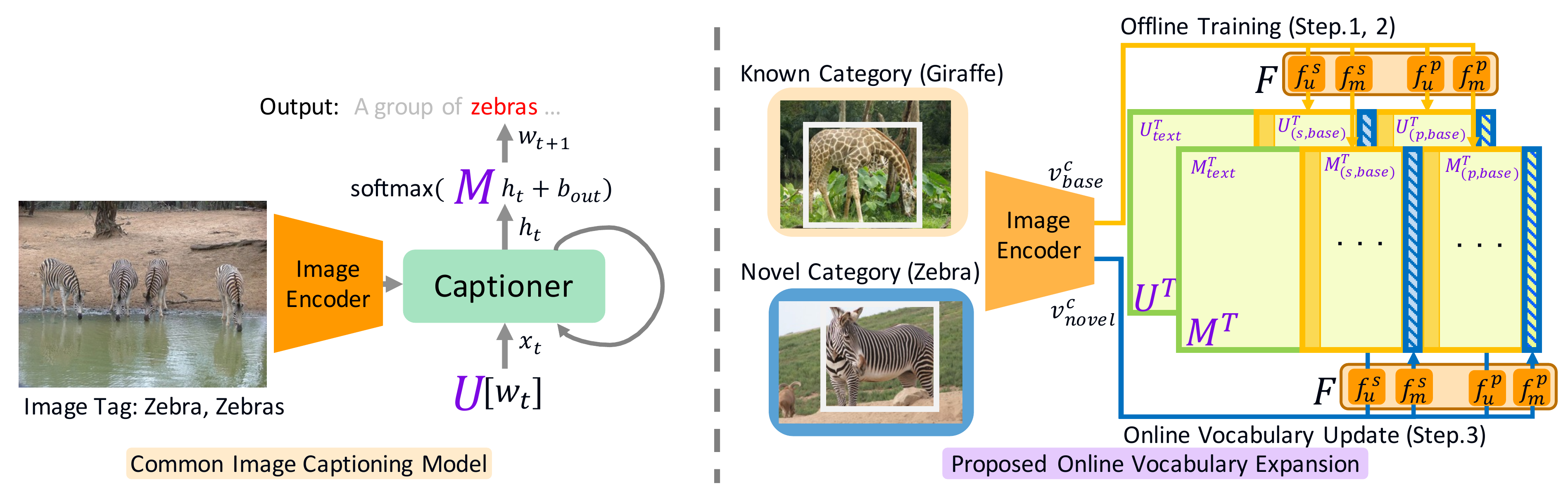}
  \caption{The proposed model is shown. The figure on the left is a commonly used image-captioning model. A module that generates words recursively, such as an LSTM, is abbreviated as a captioner. In this study, we expand matrices $U$ and $M$ shown in the left figure without retraining, as depicted in the right figure. Additionally, we generate captions from images with objects that are not included in the training dataset. From the image features of the novel objects, singular and plural word embeddings of those categories are estimated for each of the matrices $U$ and $M$.
  }
\label{fig:few_arch}
\end{figure*}

We propose a method to expand vocabulary by estimating word embeddings of novel objects from a small number of images features for captioning images with novel objects. The entire system is depicted in Fig.~\ref{fig:few_arch}. In commonly used image-captioning models, sentences are generated by inputting the previous word and then predicting the next word, and the model can be simplified as shown on the left side of the figure. The captioner in the figure comprises an LSTM and an attention module, among others. Because the proposed method changes the input and output parts, it can be applied irrespective of the structure of the captioner. The proposed model is constructed using the following procedure:

\begin{quote}
\begin{itemize}
 \setlength{\parskip}{0cm} 
 \setlength{\itemsep}{0cm} 
 \item[Step.1] (Offline) Extract image feature that represents each known category.
 \item[Step.2] (Offline) For images that contain known categories, learn image captioning by learning the converter $F$, which estimates the word embeddings using the image features calculated in Step.1.
 \item[Step.3] (Online) Prepare a small number of images of novel objects, and estimate their word embeddings using the converter $F$ trained in Step.2. The vocabulary is expanded by repeating Step.3 every time a novel object is added.
\end{itemize}
\end{quote}

We now explain each step in the chronological order.

\noindent{\textbf{Step.1 (Offline).}}
The purpose of this step is to obtain the image feature $v^c_{base}$, which represents each known category. The average image feature for each category can be calculated using datasets for image classification or object detection. In this study, image features were extracted using a model that classifies known categories. We used Faster R-CNN~\cite{renNIPS15fasterrcnn}, which is a popular object detector for obtaining features from target object regions. We set $v^c_{base}$ as the average feature of the output before the last layer of the model, for each category.

\noindent{\textbf{Step.2 (Offline).}}
In this step, an image-captioning model is learned while learning a converter $F$ that estimates word embeddings from the image feature $v^c_{base}$. By learning the converter $F$ under the objective function of image captioning, the model learns to explain images by estimating the property of known objects using their image features. In general image captioning, all the features of words are learned from random values. However, in our method, the features of the words or phrases corresponding to the known categories are estimated using their image features. For example, if ``teddy bear(s)'' is in the known categories and it is found in the annotated sentences, the phrase embedding estimated using the image feature of ``teddy bear'' is used. Word embeddings are required in two matrices, namely, the input matrix $U\in\mathbb{R}^{V\times d1}$ and the output matrix $M\in\mathbb{R}^{V \times d2}$ (where $d1$ and $d2$ denote each feature dimension and $V$ denotes the number of words in the vocabulary). Because we must consider both the singular and plural forms for each word, the converter comprises four functions, namely, $F=\{f^{s}_{u},f^{p}_{u},f^{s}_{m},f^{p}_{m}\}$. In this study, all these functions are independent linear layers and are given as follows:

\begin{eqnarray}
u^{c}_{(s,base)} &=& f^{s}_{u}(v^c_{base})\\
u^{c}_{(p,base)} &=& f^{p}_{u}(v^c_{base})\\
m^{c}_{(s,base)} &=& f^{s}_{m}(v^c_{base})\\
m^{c}_{(p,base)} &=& f^{p}_{m}(v^c_{base})
\label{eqn:gen_new_w}
\end{eqnarray}

$U_{(s,base)}$ and $U_{(p,base)}$ denote the concatenated word embeddings of each category $u^{c}_{(s,base)}$ and $u^{c}_{(p,base)}$, respectively, and the word embeddings $U_{text}$ are trained from random values. The feature $x_t$ of the input word $w_t$ to the captioner in Fig.~\ref{fig:few_arch} is calculated as follows:

\begin{eqnarray}
U &=& [U_{text};U_{(s,base)};U_{(p,base)}] \\
x_t &=& U[w_t] 
\end{eqnarray}

Similar to the input matrix $U$, the output matrix $M$ is calculated by concatenating $M_{(s,base)}, M_{(p,base)}$, and $M_{text}$. All the bias values in $b_{out}$, which indicates the ease of occurrence of each word, is learned from random values to learn from annotated sentences. The following operation is performed on the output $h_t$ from the captioner in Fig.~\ref{fig:few_arch}, and the next-word probability $p(w_t|w_1,\cdots,w_{t-1},I)$ is the output. One has the following:

\begin{eqnarray}
M &=& [M_{text};M_{(s,base)};M_{(p,base)}] \\
b_{out} &=& [b_{text},b_{(s,base)},b_{(p,base)}] \\
p(w_t|w_1, \cdots ,w_{t-1}, I) &=& {\it {\rm Softmax}}(Mh_t+b_{out})
\end{eqnarray}

The parameter $\theta$ in the caption generator and the converter $F$ is learned by minimizing the negative log-likelihood as follows:

\begin{eqnarray}
L(\theta) = -\sum_t \log p(w_t|w_1, \cdots ,w_{t-1}, I)
\label{eqn:log_likeli}
\end{eqnarray}

\noindent{\textbf{Step.3 (Online).}}
In this step, the vocabulary is expanded by providing a small number of images of novel objects. First, $v^c_{novel}$ is calculated using the same image-feature extractor as used in Step.1. Next, applying $F=\{f^{s}_{u},f^{p}_{u},f^{s}_{m}, f^{p}_{m}\}$ trained in Step.2 to the image feature of novel object $v^c_{novel}$ in combination with $v^c_{base}$, we can obtain $U_{(s,base+novel)}$, $U_{(p,base+novel)}$, $M_{(s,base+novel)}$, and $M_{(p,base+novel)}$, and the vocabulary can be expanded using the equation shown below. The word embeddings of the novel categories can be obtained without retraining, and the required function (B) can be satisfied. The terms $b_{(s,novel)}$ and $b_{(p, novel)}$ corresponding to the novel categories are explained in Sec.~\ref{sec:cap}.

\begin{eqnarray}
U &=& [U_{text};U_{(s,base+novel)};U_{(p,base+novel)}] \\
M &=& [M_{text};M_{(s,base+novel)};M_{(p,base+novel)}] \\
b_{out} &=& [b_{out},b_{(s,base+novel)},b_{(p,base+novel)}] 
\end{eqnarray}

\subsection{Image Captioning with Expanded Vocabulary}
\label{sec:cap}

Finally, we introduce a method of generating sentences for images with novel objects based on the expanded vocabulary, using the method described in Sec.~\ref{sec:ex}. As previously mentioned, in this study, we focus on $P(S|I,C_{novel})$, i.e., the image tags of novel objects are provided. Generally, the model is not expected to output words that were not generated during training. Furthermore, CBS~\cite{Anderson_2017} was adopted to effectively utilize image tags. The method performs beam search so that the sentences to be generated include image tags as constraints $C$, using a finite-state machine that can transition to the next state while outputting a constraint word. For example, multiple constraints can be set by giving $C$=\{$C1$:\{desk,desks\},$C2$:\{chair\}\}, and the sentence to be generated is searched for the condition that includes chair and desk or only desks. In this study, multi-word phrases such as ``hot dog'' are generated as a group without dividing them into words; therefore, if the number of constraints is $n$, the state number would be $2^n$. Using this algorithm, it is possible to generate sentences with the names of novel objects. However, we cannot still satisfy the required function (C) because the names of novel objects can be included even if an image contains only known objects. In addition, there is a problem that the same word is repeated when the bias value is set as relatively large experimentally. To simultaneously solve these problems, we set $b_{(s, novel)}$ and $b_{(p, novel)}$ of $b_ {out}$, which correspond to the novel objects, to sufficiently small values.

\section{Experiments}
First, we explain the datasets used and experimental settings. Next, we introduce the results for each dataset.

\subsection{Datasets and Experimental Settings}

\noindent{\textbf{Datasets.}}
We conducted experiments using two datasets. The first dataset is called the Held-out MSCOCO dataset~\cite{Hendricks_2016}, which is created by dividing MSCOCO into known and novel categories. This experiment aims to verify the effectiveness of the proposed method for each category using commonly used benchmarks. The second dataset is called the nocaps dataset~\cite{Agrawal_2019}, which is based on a large-scale object-detection dataset, Open Image dataset.
This experiment aims to verify the usefulness of the proposed method for more diverse objects. The Held-out MSCOCO dataset is constructed by excluding images with sentences that contain any of the eight classes, namely, bottle, bus, couch, microwave, pizza, racket, suitcase, and zebra, from the training data; however, the validation/test data include these objects. The nocaps dataset is constructed using 513 categories out of 600 categories of Open Image, where 119 categories are in-domain, 394 out-domain, and the images that are included in both the categories are set as near-domain. The validation data comprises 4500 images. In this study, because the train/validation data were divided in MSCOCO, we used the validation data in the nocaps dataset as the test data.

\noindent{\textbf{Experimental Settings.}}
Our method can be used by replacing the features of words used in a normal image-captioning model with those used in the proposed method. In this study, we performed an experiment based on a model called Updown~\cite{Anderson2018}, which offers high image-captioning performance. Although Updown~\cite{Anderson2018} used Faster R-CNN as an image-feature extractor whose output is fed into the captioner, we used VGG-16 pretrained on ImageNet~\cite{imagenet_cvpr09} in the Held-out MSCOCO dataset experiment to align the conditions with the previous researches. Updown with VGG features is referred to as Baseline. Baseline with the proposed method is referred to as Baseline + vis2w. The Faster R-CNN used for word embedding estimation in this study was trained to detect 72 classes for the experiment of the Held-out MSCOCO dataset and 80 for the experiment of the nocaps dataset. For Faster R-CNN, we used the open-source implementation~\cite{jjfaster2rcnn}. For each dataset, we first performed ablation studies upon obtaining sufficient number (approximately 1000) of images for each novel object, and then we showed the performance when the number of available images was changed. We used a beam size of five.

\subsection{Results on the Held-out MSCOCO Dataset}

\begin{figure*}[t]
\centering
  \includegraphics[width=\linewidth]{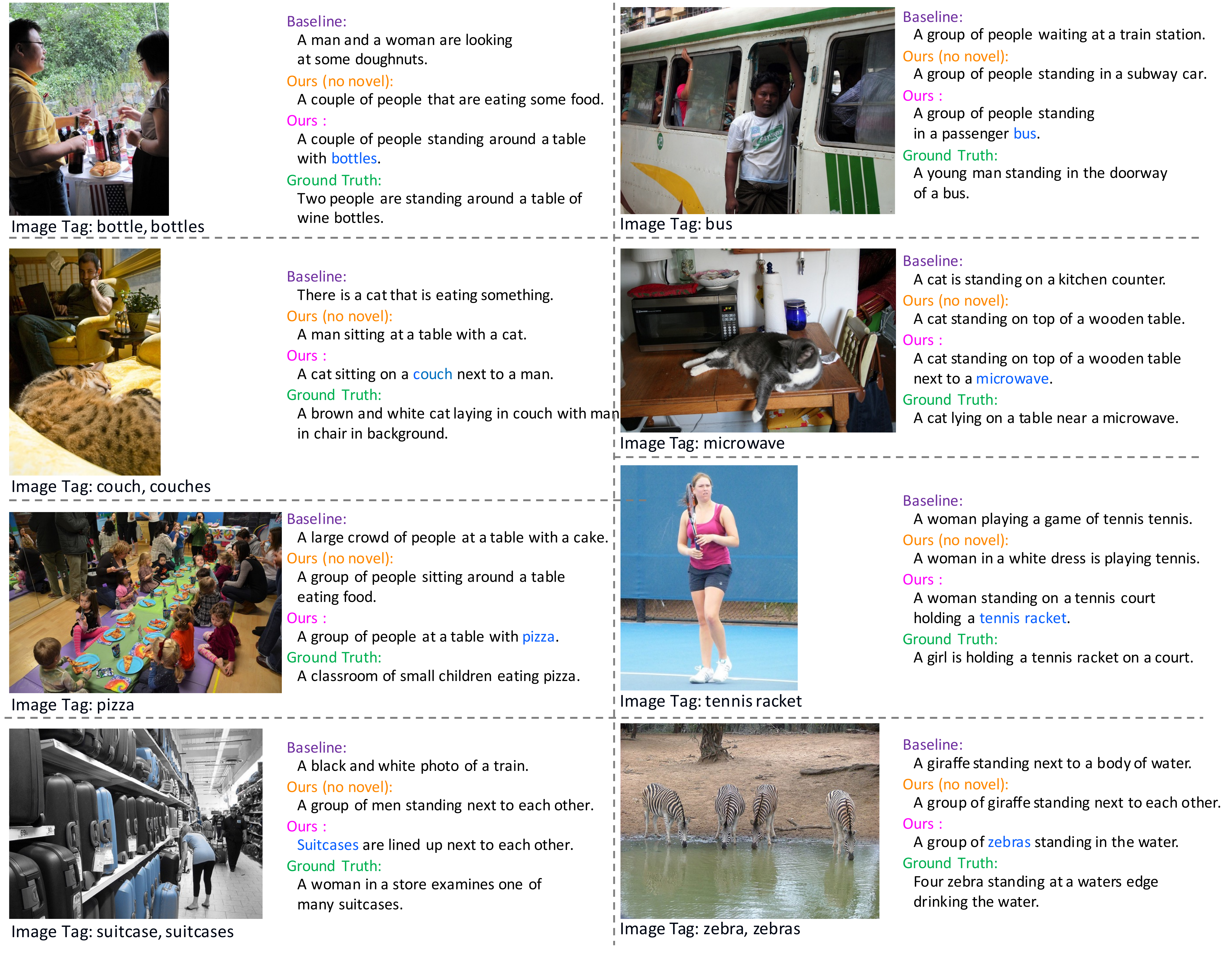}
  \caption{Comparison of sentences generated from images with novel objects in the Held-out MSCOCO dataset. The words/phrases shown in blue are the names of novel objects.
  }
\label{fig:few_gen}
\end{figure*}

\noindent{\textbf{Qualitative Results.}}
Fig.~\ref{fig:few_gen} shows the sentences generated from images with novel objects. Image tags are given when there are annotations of novel objects, and plural form is added only when there are multiple objects of the same categories so that the model can select the word to be used while generating sentences. For example, if there is one zebra in an image, the constraint is $C$=\{zebra\}; however, if the image contains two or more zebras, the constraint is $C$=\{zebra, zebras\}. By estimating the word embeddings of the novel objects from their image features, our model generates sentences including novel objects by effectively using image tags. As shown in the example of zebra in the lower right of the figure, the proposed method can generate a sentence by selecting the word to be used when both singular and plural words are given.

\noindent{\textbf{Quantitative Results.}}
First, we evaluated the sentences generated using our method using an automatic evaluation metric CIDEr~\cite{Vedantam2015}. Table~\ref{table:FMCOCO} shows the results of the ablation studies. Because the performance of Baseline + vis2w, which estimates word embeddings of known categories using image features, was close to that of Baseline, the performance of generating image descriptions consisting of known objects did not deteriorate upon using our method. For Baseline + vis2w, the performance did not change when the constraint was zero, even when estimating the word embeddings of both the singular and plural forms of the novel categories, as the bias values corresponding to the novel categories were set as sufficiently small, as described in Section ~\ref{sec:ex}. CBS was performed when the constraint number was set to 1 or more, and it improved the performance. Comparing Baseline + vis2w with the singular forms of novel objects and with both singular and plural forms of novel objects, the performance was improved, especially in the case of images with zebras. This is because zebras often form a group, and, therefore, we must consider plural forms while explaining images that contain zebras. Our method generates sentences by effectively selecting singular or plural forms.

\begin{table*}[!tb]
\setlength{\tabcolsep}{3.5pt}
\centering
\caption{Generation evaluation based on CIDEr on the Held-out MSCOCO dataset is shown. In the column of Novel, s and p means that word embeddings of singular and plural forms are estimated respectively, and s + p means that both of them are estimated. Constraint column shows the number of constraints used for CBS and if the number is zero, normal beam search is performed. Each column of eight categories represent the performance for one of the eight novel categories. Next, 8 Categories represents the performance for images containing at least one of the eight novel categories. The rightmost Test Set shows the average performance of all test data that includes both known categories and the novel categories. The proposed method especially improves the performance of novel categories.
}
\resizebox{1\columnwidth}{!}{%
\label{table:FMCOCO}
\begin{tabular}{|l|l|l||cccccccc|c|c|}\hline
Methods&Novel&Constraint&Bottle&Bus&Couch&Microwave&Pizza&Racket&Suitcase&Zebra&8 Categories&Test Set
\\ \hline
Baseline &-&0& 0.659 & 0.423 &0.680 &0.628 &0.417 &0.798 & 0.407 &0.348 &0.532 &0.836\\ \hline
Baseline + vis2w &-&0& 0.657 & 0.404 &0.721 &0.695 &0.419 &0.871 & 0.395 &0.357 &0.547 &0.834\\ \hline
Baseline + vis2w &s+p&0& 0.657 & 0.404 &0.721 &0.695 &0.419 &0.871 & 0.395 &0.357 &0.547 &0.834\\ \hline
Baseline + vis2w &s & 1 & 0.648 & 0.792 &0.969 &\bf0.984 &\bf0.866 &1.037 & 0.534 &0.748 &0.829 &0.852\\ \hline
Baseline + vis2w &s+p& 2 & \bf0.675 & 0.786 &0.955 &0.964 &0.823 &\bf1.041 & 0.559 &\bf0.841 &0.834 &0.852\\ \hline
Baseline + vis2w &s+p& 1 & 0.672 & \bf0.797 &\bf0.974 &\bf0.984 &0.865 &1.039 & \bf0.562 &\bf0.841 &\bf0.847 &\bf0.855\\ \hline
\end{tabular}
}
\end{table*}

\begin{table*}[!tb]
\setlength{\tabcolsep}{3.5pt}
\centering
\caption{
The change in the performance is shown when the number of images of novel objects used to generate word embeddings is changed for the Held-out MSCOCO dataset. The leftmost column shows the number of the annotations used. Other columns are the same as in Table~\ref{table:FMCOCO}. The average and error ranges are shown for 50 patterns randomly selected from the annotations of the novel categories included in the train split. The more are the images used, the higher becomes the performance, and the performance achieved using 50 instances is almost similar to that achieved upon using 1000 instances.
}
\resizebox{1\columnwidth}{!}{%
\label{table:few_num}
\begin{tabular}{|l|cccccccc|c|c|}\hline
\# of annotations&Bottle&Bus&Couch&Microwave&Pizza&Racket&Suitcase&Zebra&8 Categories&Test Set
\\ \hline
0 (w/o novel) & 0.657 & 0.404 &0.721 &0.695 &0.419 &0.871 & 0.395 &0.357 &0.547 &0.834\\ \hline \hline

1 &0.686 $\pm$ 0.011&
0.766 $\pm$ 0.008&
0.904 $\pm$ 0.012&
0.950 $\pm$ 0.011&
0.779 $\pm$ 0.015&
1.018 $\pm$ 0.013&
0.541 $\pm$ 0.006&
0.820 $\pm$ 0.014&
0.809 $\pm$ 0.004&
0.848 $\pm$ 0.001\\ \hline

5 &0.695 $\pm$ 0.006&
0.790 $\pm$ 0.005&
0.946 $\pm$ 0.008&
0.973 $\pm$ 0.008&
0.851 $\pm$ 0.006&
1.035 $\pm$ 0.007&
0.560 $\pm$ 0.004&
0.837 $\pm$ 0.009&
0.840 $\pm$ 0.002&
0.854 $\pm$ 0.001\\ \hline

10 & 0.694 $\pm$ 0.004&
0.793 $\pm$ 0.004&
0.955 $\pm$ 0.007&
0.973 $\pm$ 0.006&
0.854 $\pm$ 0.005&
1.037 $\pm$ 0.005&
0.564 $\pm$ 0.003&
0.847 $\pm$ 0.006&
0.844 $\pm$ 0.002&
0.854 $\pm$ 0.000\\ \hline

50 & 0.684 $\pm$ 0.003&
0.797 $\pm$ 0.001&
0.962 $\pm$ 0.004&
0.983 $\pm$ 0.004&
0.857 $\pm$ 0.003&
1.035 $\pm$ 0.002&
0.561 $\pm$ 0.002&
0.852 $\pm$ 0.004&
0.846 $\pm$ 0.001&
0.855 $\pm$ 0.000\\  
\hline\hline

1000 & 0.672 & 0.797 &0.974 &0.984 &0.865 &1.039 & 0.562 &0.841 &0.847 &0.855\\ \hline
\end{tabular}
}
\end{table*}

Next, we conducted an experiment to examine the change in the sentence-generation performance upon changing the number of annotations. The results are presented in Table~\ref{table:few_num}. The proposed method performed even with one image, and the sentence-generation performance improved upon collecting more images of novel categories. The performance achieved using 50 annotations was almost the same as that achieved using 1000 annotations. This result indicates the following. First, we can generate better sentences than sentences that simply contain the names of novel objects using better word embeddings estimated by multiple images. Second, the proposed method, which estimates the word embeddings of novel objects using their image features, can perform even if only a few images of those novel categories are provided.

\begin{table*}[!tb]
\setlength{\tabcolsep}{3.5pt}
\centering
\caption{
Comparison between results of our method and those of existing methods. Notably, different methods require different annotations for novel objects. In our setting, novel categories are added from a small number of images without any retraining. We show the results of average and error ranges when 50 patterns are randomly selected from the train split.
}
\resizebox{0.6\columnwidth}{!}{%
\label{table:few_compare}
\begin{tabular}{|l||ccc|}\hline
Methods&SPICE&METEOR&CIDEr
\\ \hline
LRCN~\cite{Donahue2015} &-& 0.193 &- \\ \hline
\multicolumn{4}{c}{Images and texts are needed}\\
\hline
DCC~\cite{Hendricks_2016}  &0.134& 0.210 &0.591 \\ \hline
NOC~\cite{Venugopalan_2017}  &-& 0.207 &- \\ \hline
Base+T4~\cite{Anderson_2017} &0.159& 0.233 &0.779 \\ \hline
LKGA-CGM~\cite{Mogadala2017}  &0.146& 0.222 &- \\ \hline
LSTM-C~\cite{Yao_2017} &-& 0.230 &- \\ \hline
LSTM-P~\cite{Li_2019}  &0.166& 0.234 &0.883 \\ \hline
\multicolumn{4}{c}{Texts only}\\
\hline
ZSC~\cite{Demirel_2019} &0.142& 0.219 &- \\ \hline
\multicolumn{4}{c}{Only images for training a detector}\\
\hline
NBT~\cite{Lu2018}  &0.157& 0.228 &0.770 \\ \hline
DNOC~\cite{Wu_2018}  &-& 0.216 &- \\\hline
CRN~\cite{Feng_2020} &-& 0.213 &- \\\hline
\multicolumn{4}{c}{Only images are provided online and image tags are given}\\
\hline
Baseline + vis2w  (0 image)&0.115&0.196&0.522 \\ \hline
Baseline + vis2w  (1 image)&0.165 $\pm$ 0.001&0.222 $\pm$ 0.001&0.780 $\pm$ 0.005\\ \hline
Baseline + vis2w  (5 image)&0.168 $\pm$ 0.001&0.225 $\pm$ 0.001&0.802 $\pm$ 0.003 \\ \hline
Baseline + vis2w  (10 image)&0.168 $\pm$ 0.000&0.225 $\pm$ 0.000&0.803 $\pm$ 0.002 \\ \hline
Baseline + vis2w  (50 image)&0.169 $\pm$ 0.000&0.225 $\pm$ 0.000&0.805 $\pm$ 0.001 \\ \hline
Baseline + vis2w  (1000 image)&0.169&0.226&0.808 \\ \hline
\end{tabular}
}
\end{table*}

Finally, we compared our method to those in existing studies, to show that similar performance could be achieved by our settings, as shown in Table~\ref{table:few_compare}. We used a beam size of one in this experiment. Our setting is different from existing studies as follows. First, our setting is disadvantageous in that novel categories are added from a small number of images without retraining, and resources are limited compared to the existing ones. On the other hand, our setting is advantageous in that we used GT image tags, and the search width of CBS with beam size of one is approximately twice that of greedy decoding. When compared under these differences, our method performs similar to the existing ones.

\subsection{Results on the Nocaps Dataset}

\begin{figure*}[htbp]
\centering
  \includegraphics[width=\linewidth]{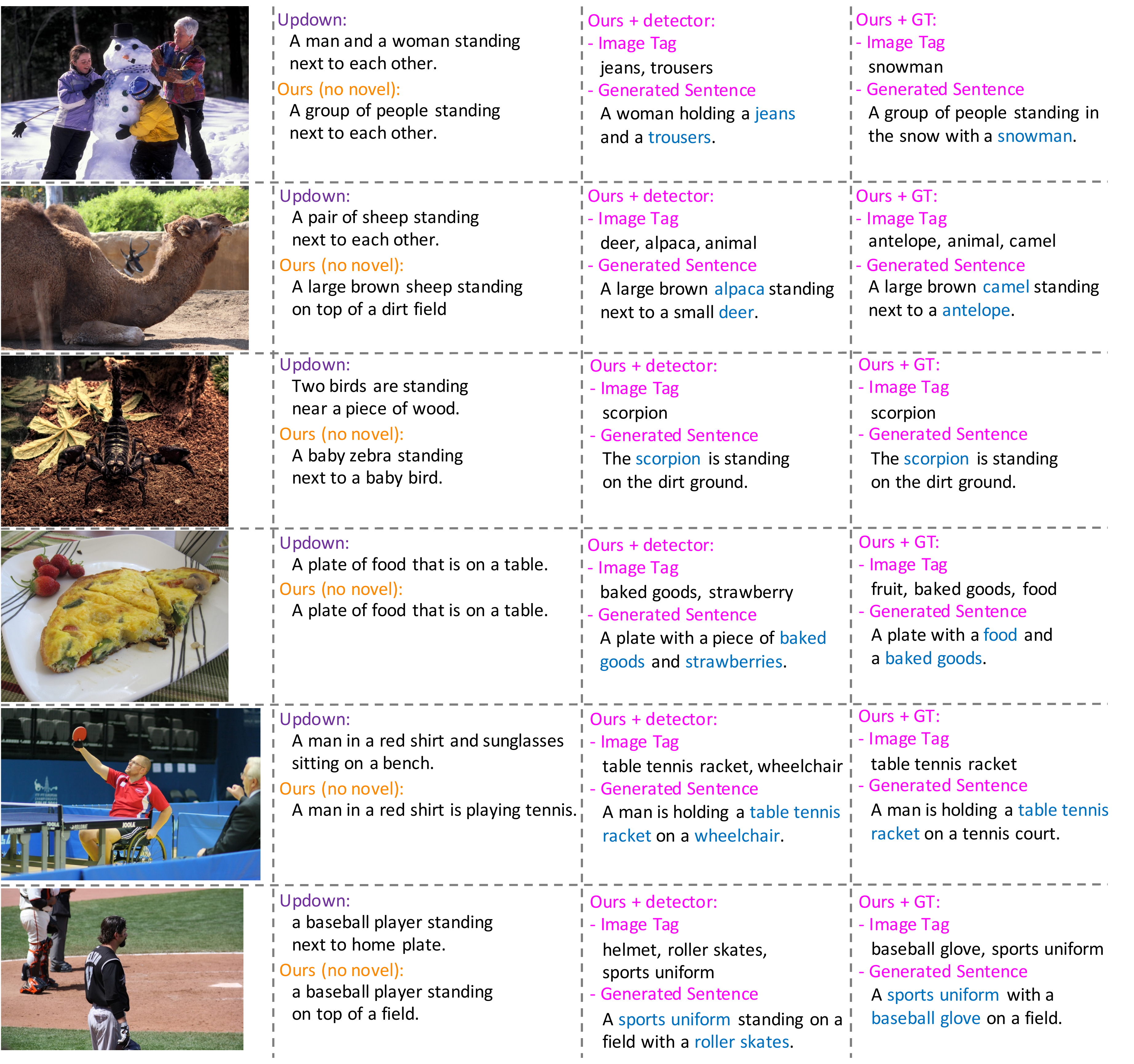}
  \caption{Comparison of sentences generated from images with novel objects on the nocaps dataset. From the left column, input images, generated sentences by Updown and Updown with the proposed word embeddings for known categories, generated sentences using the proposed method with image tags from a detector used in \cite{Agrawal_2019}, and generated sentences using the proposed method with image tags from GT. We provide the singular and plural forms as image tags as in \cite{Agrawal_2019}, and the plural forms are abbreviated. The first and second rows are the examples of success using tags from GT. The third row shows the case wherein the same image tags are obtained from the detector and GT. The next two rows show the examples of success using tags from the detector. The bottom example is of a failure case of the proposed method. The words or phrases depicted in blue correspond to those used as constraints from image tags.
  }
\label{fig:few_no_gen}
\end{figure*}

\noindent{\textbf{Qualitative Results.}}
The sentences generated using the nocaps dataset are shown in Fig.~\ref{fig:few_no_gen}. We experimented on two types of image tags, one from the detector used in \cite{Agrawal_2019} and the other from GT annotations. Upon using image tags from GT in the rightmost column, we performed the same filtering as in \cite{Agrawal_2019} by sorting objects in the order of their areas. As can be seen from the top two rows, the proposed method can generate sentences that include novel objects under the correct image tags. In the proposed method, as seen from the second row from the top, because word embeddings are estimated using image features in this study, ``an antelope'' is output incorrectly as ``a antelope''; however, this kind of errors can be easily modified via rule-based post-processing. However, the method using tags from the detector either uses unnecessary tags for explanation, which are often detected, as shown in the top row, or generates sentences using tags that are incorrectly recognized, as shown in the second row from the top. The third row from the top shows the case wherein the same tag is obtained by the detector and GT, and both of them output the correct sentence by effectively using the image tag. The fourth and fifth rows show the case wherein the detector is able to output better image tags than those by GT because of the missing annotations, and the sentences that use image tags from the detector describe images more faithfully. The bottom row shows a failure example of the proposed method. Because of the visual similarity, the word embedding of ``sports uniform'' was estimated to be close to that of the person.

\noindent{\textbf{Quantitative Results.}}
First, we evaluated the sentences generated using our method based on automatic evaluation metrics, namely, SPICE~\cite{Anderson_2016} and CIDEr. Table~\ref{table:comp_eval} shows the results of the ablation studies. We used two types of image tags, one from the detector used in \cite{Agrawal_2019} and the other from GT. 
In \cite{Agrawal_2019}, the performance of generating sentences that include words that have never been generated during the training was improved upon using ELMO~\cite{Peters2018}, which can perform in a context wider than Glove. However, our proposed method could not reach the performance of Updown + ELMO, as \cite{Agrawal_2019} used the word embeddings pretrained on a large-scale text-corpus that included novel objects. In our setting, there were no images or texts associated with novel objects during the training, and no training was required to add novel categories. Comparing Updown with Updown + vis2w or with Updown + vis2w + GT, the performance of the proposed method improves as the domain moves away from the in-domain. Comparing Updown + vis2w + GT with the singular forms of novel objects and with both singular and plural forms of novel object, the performance in the latter case is better, and the proposed model can generate sentences by appropriately selecting singular or plural forms of novel objects. Contrary to the experiment on the Held-out MSCOCO dataset, Updown + vis2w + GT with two constraints is better than that with one constraint, as there are more types of novel categories in the nocaps dataset, and two or more novel objects often need to be described.

\begin{table*}[!tb]
\scriptsize
\centering
\caption{Sentence-generation evaluation based on SPICE and CIDEr on the nocaps dataset is shown. + detector and + GT denote the source of image tags. Note that Updown + ELMO and Updown + Elmo + GT are trained under the setting that large amount of sentences that include novel objects can be used during training, and our setting is different because novel categories are added without training. The performance is mainly improved in the out-domain using the proposed method.
}
\resizebox{0.8\columnwidth}{!}{%
\begin{tabular}{| l | l | l || c | c | c | c | c | c| c| c|}
\hline
&&&\multicolumn{2}{c}{in-domain} &\multicolumn{2}{c}{near-domain} &\multicolumn{2}{c|}{out-domain} &\multicolumn{2}{c|}{Overall}\\
\cline{4-11}
Methods& Novel&Constraint& CIDEr &SPICE&CIDEr&SPICE&CIDEr&SPICE&CIDEr&SPICE\\
\hline
Updown~\cite{Agrawal_2019} 	&-& 0&78.1& 11.6 &57.7 &10.3 &31.3 &8.3 &55.3& 10.1 \\
Updown + ELMO + detector~\cite{Agrawal_2019}				&s+p&2&79.3& 12.4& 73.8& 11.4& 71.7 &9.9& 74.3 &11.2\\
Updown + ELMO + GT~\cite{Agrawal_2019}  &s+p&2&84.2& 12.6& 82.1& 11.9& 86.7 &10.6& 83.3 &11.8\\
\hline\hline
Updown + vis2w	&-&0	& 74.72 & 11.46 & 54.5 & 10.06 & 26.61 & 7.72 & 51.74 & 9.88 \\
Updown + vis2w + detector	&s+p&2& 71.9 & 11.0 & 68.5 & 10.78 &  66.08& 9.82 &68.5 & 10.62 \\
Updown + vis2w + GT 	&s&2& 79.03 & 11.57 & 74.4& 11.05 &  \bf78.57& 10.47 & 75.91 & 11.01 \\
Updown + vis2w + GT 	&s+p&1& \bf80.05 & \bf11.64 & 72.69 & 11.02 &  68.97& 10.21 & 73.0 & 10.95 \\
Updown + vis2w + GT 	&s+p&2& 79.53 & 11.54 & \bf74.67 & \bf11.11 &  78.21& \bf10.67 & \bf76.09 & \bf11.08 \\
\hline
\end{tabular}}
\label{table:comp_eval}
\end{table*}

\begin{table*}[!tb]
\scriptsize
\centering
\caption{
The change in the performance is shown when the number of the images of novel objects used to generate word embeddings is changed on the nocaps dataset. Notably, \# of annotations$\ast$ indicates the maximum number of annotations because the number of the training data are less than that described for some categories. The average and error ranges are shown for 50 random patterns selected from the annotations of the novel categories included in the train split. The proposed method improves the out-domain performance even with one image.
}
\resizebox{\columnwidth}{!}{%
\begin{tabular}{| l || c | c | c | c | c | c| c| c|}
\hline
&\multicolumn{2}{c}{in-domain} &\multicolumn{2}{c}{near-domain} &\multicolumn{2}{c|}{out-domain} &\multicolumn{2}{c|}{Overall}\\
\hline
\# of annotations$\ast$ & CIDEr &SPICE&CIDEr&SPICE&CIDEr&SPICE&CIDEr&SPICE\\
\hline
0 (w/o novel) 	&78.1& 11.6 &57.7 &10.3 &31.3 &8.3 &55.3& 10.1 \\
\hline\hline
1 	&78.123 $\pm$ 0.201& 11.579 $\pm$ 0.035 &72.283 $\pm$ 0.140 &11.015 $\pm$ 0.022 &74.406 $\pm$ 0.288 &10.370 $\pm$ 0.036 &73.555 $\pm$ 0.119& 10.967 $\pm$ 0.019 \\
5 	&79.169 $\pm$ 0.199& 11.565 $\pm$ 0.028 &74.103 $\pm$ 0.101 &11.074 $\pm$ 0.014 &77.124 $\pm$ 0.163 &10.523 $\pm$ 0.025&75.446 $\pm$ 0.086& 11.034 $\pm$ 0.012 \\
10 	&79.247 $\pm$ 0.124& 11.553 $\pm$ 0.016 &74.327 $\pm$ 0.085 &11.083 $\pm$ 0.012 &77.592 $\pm$ 0.157 &10.564 $\pm$ 0.024&75.699 $\pm$ 0.072& 11.047 $\pm$ 0.010 \\
50	&79.347 $\pm$ 0.094& 11.538 $\pm$ 0.011 &74.566 $\pm$ 0.062 &11.095 $\pm$ 0.009 &78.203 $\pm$ 0.111 &10.613 $\pm$ 0.016 &75.992 $\pm$ 0.047& 11.062 $\pm$ 0.007 \\
\hline\hline
1000 	& 79.53 & 11.54 & 74.67 & 11.11 &  78.21& 10.67 & 76.09 & 11.08 \\
\hline
\end{tabular}}
\label{table:nocaps_num_eval}
\end{table*}

Next, we conducted an experiment to examine the change in the sentence-generation performance upon is changing the number of annotations. The results are presented in Table~\ref{table:nocaps_num_eval}. The performance increases as the domain moves away from the in-domain. The proposed method can perform with a small number of images of novel objects, even when there is a single image for each category. As the number of annotations increased, the sentence-generation performance also increased in the same way as in Table~\ref{table:few_num}.

\section{Conclusions}
In this research, we examined captioning images that include novel objects with tags by using small amount of data without retraining. Considering that humans estimate the property of novel objects by associating them with known categories based on their visual information, we proposed a method of generating captions from images that include novel objects with their word embeddings estimated using a small number of images. The method can be applied to general image-captioning models while reducing data-collection and retraining costs.

\section*{Acknowledgment}
\addcontentsline{toc}{section}{Acknowledgment}
This work was partially supported by JST CREST Grant Number JPMJCR1403, and partially supported by JSPS KAKENHI Grant Number JP19H01115. We would like to thank Atsuhiro Noguchi, Hiroaki Yamane, Yusuke Mukuta and James Borg for helpful discussions.

\bibliographystyle{splncs04}
\bibliography{egbib}
\end{document}